\documentclass[letterpaper, 10 pt, conference]{ieeeconf}  % Comment this line out if you need a4paper

\IEEEoverridecommandlockouts                              % This command is only needed if 
\usepackage{times}
\usepackage{graphics} % for pdf, bitmapped graphics files
\usepackage{graphicx}
\usepackage{subfigure}
\usepackage{amsmath,amssymb,amsopn,amstext,amsfonts}
\usepackage{cancel}
\usepackage[space]{cite}
\usepackage{pdfsync}
\usepackage{balance}
\usepackage{color}
\usepackage{mathtools}
\usepackage{bm}

\usepackage{diagbox}
\usepackage{float}
\usepackage{epstopdf}
\usepackage{pifont}
\usepackage{fixltx2e}
\usepackage{amsmath}
\usepackage{multirow}
\usepackage{url}
\usepackage{verbatim}
\usepackage{caption}
\usepackage{adjustbox}
\usepackage{booktabs}
\usepackage{threeparttable}
\usepackage{makecell}
\usepackage{tabularx}
\usepackage{arydshln}
\usepackage[normalem]{ulem}

\usepackage{algorithm}
\usepackage{ifthen}

\usepackage{algorithmic}

\usepackage{array}
\usepackage{textcomp}
\usepackage{stfloats}

\newcommand{\etal}{\textit{et al}.}

\makeatletter
\let\NAT@parse\undefined
\makeatother 
\usepackage[linkcolor=red,citecolor=green,urlcolor=red,colorlinks=true]{hyperref}
\title{\LARGE \bf Real-time Motion Segmentation with Event-based Normal Flow}
\author{Sheng Zhong$^{\ast}$, Zhongyang Ren$^{\ast}$, Xiya Zhu, Dehao Yuan, Cornelia Ferm\"{u}ller, Yi Zhou% <-this % stops a space
\thanks{Sheng Zhong, Xiya Zhu and Yi Zhou are with the Neuromorphic Automation and Intelligence Lab (NAIL) at School of Artificial Intelligence and Robotics,  Hunan University, Changsha, China.}
\thanks{Zhongyang Ren conducted this work while he was at the Neuromorphic Automation and Intelligence Lab (NAIL), School of Artificial Intelligence and Robotics, Hunan University, Changsha, China.}
\thanks{Dehao Yuan is with Capital One.}
\thanks{Cornelia Ferm\"{u}ller is with University of Maryland, College Park.}
\thanks{$\ast$ denotes equal contribution.}
\thanks{Corresponding author: Yi Zhou. Email: {\tt\small eeyzhou@hnu.edu.cn}.}
\thanks{This work was supported by the National Key Research and Development Project of China under Grant 2023YFB4706600.
}
}

\def\bfn{\mathbf{n}}

\def\bfu{\mathbf{u}}

\def\bfx{\mathbf{x}}

\def\bparams{\boldsymbol{\theta}}

\definecolor{light-gray}{gray}{0.5}
\newcommand\gframe[1]{{\color{light-gray}\frame{#1}}}
\begin{document}
\maketitle
\thispagestyle{empty}
\pagestyle{empty}

\begin{abstract}
Event-based cameras are bio-inspired sensors with pixels that independently and asynchronously respond to brightness changes at microsecond resolution, offering the potential to handle visual tasks in challenging scenarios.
However, due to the sparse information content in individual events, directly processing the raw event data to solve vision tasks is highly inefficient, which severely limits the applicability of state-of-the-art methods in real-time tasks, such as motion segmentation—a fundamental task for dynamic scene understanding.
Incorporating normal flow as an intermediate representation to compress motion information from event clusters within a localized region provides a more effective solution. 
In this work, we propose a normal flow-based motion segmentation framework for event-based vision.
% Building upon VecKM\_flow, we significantly improve the network's runtime speed by applying reasonable downsampling, without compromising the accuracy of normal flow estimation. 
% Building upon VecKM\_flow, we refine the network architecture to achieve real-time computational efficiency without compromising the accuracy of normal flow estimation.
Leveraging the dense normal flow directly learned from event neighborhoods as input, we formulate the motion segmentation task as an energy minimization problem solved via graph cuts, and optimize it iteratively with normal flow clustering and motion model fitting. 
By using a normal flow-based motion model initialization and fitting method, the proposed system is able to efficiently estimate the motion models of independently moving objects with only a limited number of candidate models, which significantly reduces the computational complexity and ensures real-time performance, achieving nearly a 800$\times$ speedup in comparison to the open-source state-of-the-art method.
% This substantial reduction in computational complexity ensures that our system operates in real time.
Extensive evaluations on multiple public datasets fully demonstrate the accuracy and efficiency of our framework.
% Additionally, we tested the normal flow on tasks such as angular velocity estimation and collision time detection, proving the generality of the proposed network. 
% Additionally, we also test the normal flow on tasks such as angular velocity estimation and time-to-collision estimation, demonstrating the versatility of our network in downstream applications.
Our code is released at \href{https://github.com/NAIL-HNU/EvMotionSeg}{https://github.com/NAIL-HNU/EvMotionSeg} to facilitate further research in this field.
\end{abstract}
% \input{chapters/01_intro}
% \section{System Overview}
% \label{sec:system overview}

% \input{floats/fig_system_overview}

% % A brief introduction of each module.
% \joey{
% Since the proposed ESVO2 system is built on top of the original ESVO, we 
% review the ESVO system and introduce
% the new components }

% \joey{
% In this section, we revisit the }
% \section*{Multimedia Material}
% \noindent Supplemental video: {\small \url{https://youtu.be/xxx}}\\
% Code: {\small\url{https://github.com/NAIL-HNU/Stereo-DEVO.git}}

\section{Introduction}
\label{sec: introduction}
Neuromorphic visual sensors, also known as event cameras, are bio-inspired devices that asynchronously capture pixel-level brightness changes with microsecond temporal resolution.
Compared to conventional frame-based cameras, they offer higher dynamic range~\cite{Lichtsteiner08ssc} and lower latency, making them ideal candidates for robotic perception tasks, such as optical flow estimation\cite{akolkar2020pami, kepple2020eccv}, Simultaneous Localization and Mapping\cite{Rosinol18ral, zhou2021esvo, klenk2023devo, niu2025esvo2}, and motion segmentation\cite{zhou2021emsgc, zhou2024icra_jstr, wan2025ijcv} under high-speed motion or challenging lighting conditions.

\begin{figure}[t]
	\centering
    % \subfigure[Result at $t-1$]{\gframe{\includegraphics[width=0.15\textwidth]{image/initialization/8_next.png}}}\hspace{1pt}
    \subfigure[Normal flow orientation map.]{\gframe{\includegraphics[width=0.234\textwidth]{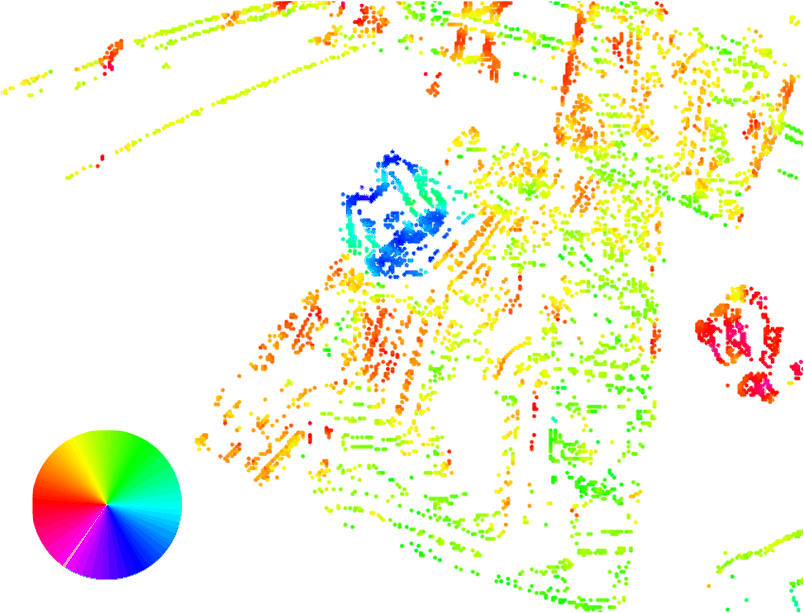}}}\hspace{5pt}
	\subfigure[Normal flow magnitude map.]{\gframe{\includegraphics[width=0.234\textwidth]{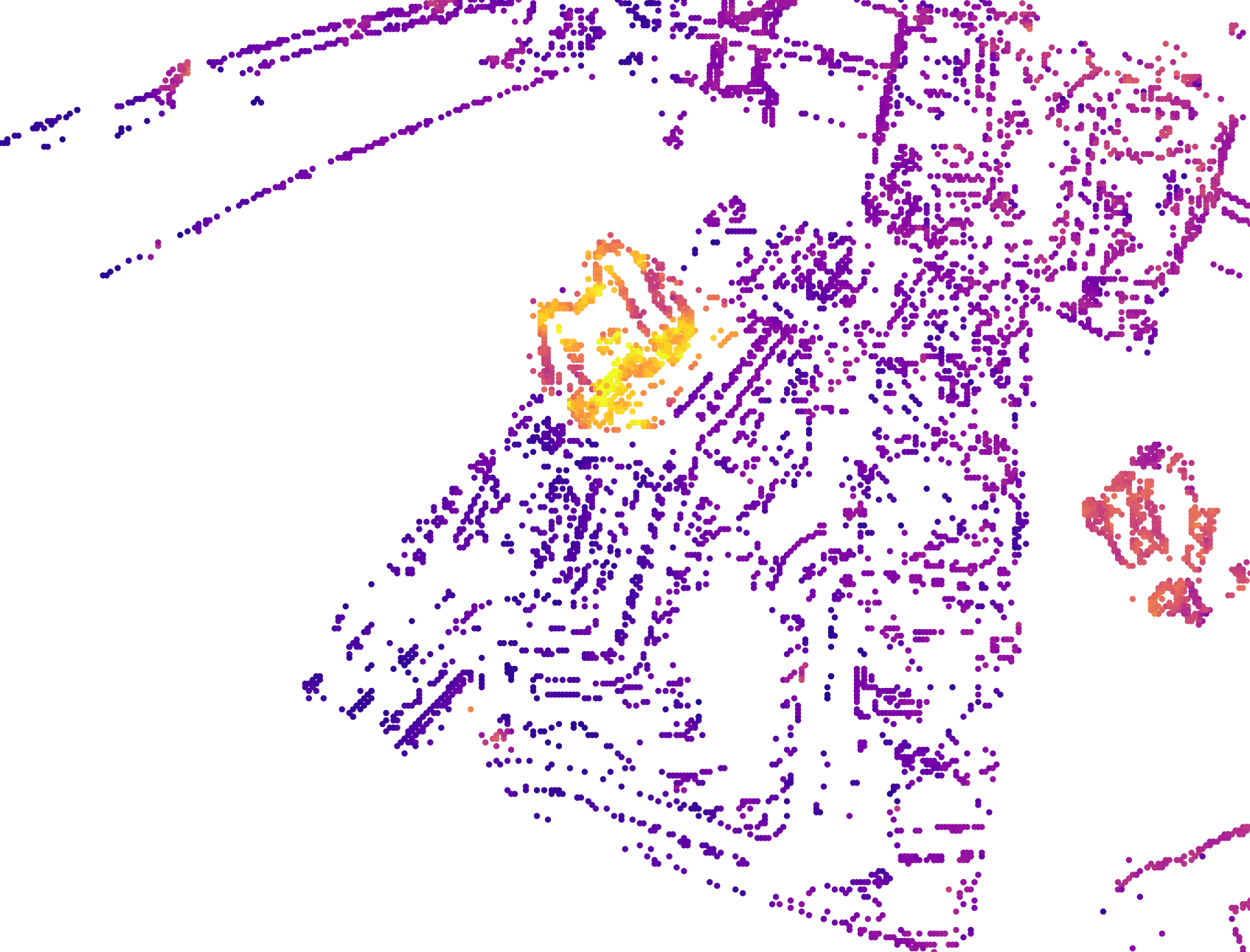}}}\\
	\subfigure[Segmentation result.]{\gframe{\includegraphics[width=0.234\textwidth]{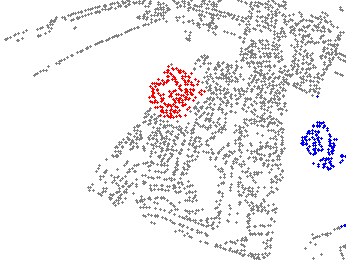}}}\hspace{5pt}
	\subfigure[Average runtime.]{\includegraphics[width=0.234\textwidth]{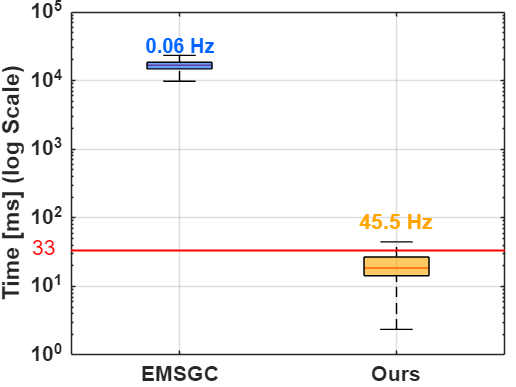}}
	\caption{The proposed system takes the normal flow generated by VecKM\_Flow~\cite{yuan2024learning} as input and performs motion segmentation in real-time based on the normal flow constraint. 
    (a) and (b) show the orientation and magnitude of the normal flow, respectively, with a circular color diagram in (a) indicating the angle-color correspondence. 
    (c) presents the motion segmentation results of our system, with different colors representing distinct motion models. 
    (d) compares the average runtime between our system and EMSGC~\cite{zhou2021emsgc} under identical setup. 
    The mean operating frequency of each method are displayed above the corresponding box, with our system achieving a speedup of nearly 800$\times$ compared to EMSGC. } 
    % There are a large number of non-edge pixels in TOS and SILC, which makes it difficult to directly sample the edge pixels, while AA accurately describes the environmental edge information and there are almost no redundant points.
 % \textcolor{red}{To revise the figure.}}
    \label{fig:eyecatcher}
    \vspace{-1.5em}
\end{figure}
Motion segmentation with event cameras involves partitioning event streams into distinct clusters, each corresponding to either the background or an independently moving object (IMO). 
The introduction of event cameras proves particularly beneficial in high-dynamic environments~\cite{gallego2020PAMI}, enabling the capture of fast movements but hardly incurring any motion blur, as is common in traditional cameras. 
However, the camera's ego-motion induces additional complexity, as it generates events across the entire image plane~\cite{glover2016event, mitrokhin2018iros, Stoffregen19iccv}, complicating the segmentation process.
The key challenge in addressing this problem is determining how to effectively assess data association corresponding to each IMO.  
One influential approach is motion compensation~\cite{gallego2018cvpr}\cite{gallego2019CVPR}, in which events are warped according to geometric models, and the goodness of fitting is evaluated using the variance of the Image of Warped Events (IWE).
The majority of the literature~\cite{mitrokhin2018iros, mitrokhin2019IROS, parameshwara2020moms, parameshwara20210, zhou2024icra_jstr} typically follows the motion compensation scheme to cluster events based on the goodness of the motion model fit, measured by various focal loss metrics.
Deep learning-based methods~\cite{mitrokhin2020learning, wan2025ijcv} have also been explored to solve this task.
Both types of methods have proven effective, but they either rely on prior knowledge of IMOs or require ground truth labels, which are not always available in real-world applications due to the lack of event data annotations.

To mitigate this dependency on such prior information, a spatio-temporal graph and Markov Random Field (MRF) are introduced in EMSGC~\cite{zhou2021emsgc}, formulating a graph-based energy function.
This approach is among the first to address event-based motion segmentation via graph-cut optimization.
% The expectation-maximization (EM) approach is then reformulated into a minimization framework using negative representation, ensuring compatibility with graph-cut algorithms.
Though effective, EMSGC~\cite{zhou2021emsgc} still has strong limitations in the computational cost of graph construction, naive initialization strategy, and iterative motion model fitting, limiting its real-time performance.

% In~\cite{mitrokhin2018iros}, global motion compensation was applied to the event stream using parametric models, and clusters of events that deviated from the motion model (residual motion) were labeled as IMOs. 
% However, such approaches encounter significant difficulties in scenarios with noise interference (e.g., poor illumination) or minimal relative motion between targets and background.

% The goal of this work is to address the limitations of the EMSGC framework by introducing normal flow,
% \textcolor{cyan}{Add explanation of normal flow here, around 2 sentences would be enough}
% which is the partial component of optical flow along the direction of image gradient.
Normal flow, the partial component of optical flow along the direction of the image gradient, provides a potential solution to these limitations.
Recently, De et al.~\cite{yuan2024learning} propose VecKM\_Flow, which is capable of directly learning dense normal flow from the event neighborhood.
Leveraging the dense normal flow as input, we propose a normal flow-based motion segmentation framework building upon EMSGC that achieves comparable segmentation accuracy while enabling real-time performance, which is critical for this class of tasks. 
% Building upon EMSGC, we propose a normal flow-based motion segmentation framework that achieves comparable segmentation accuracy while enabling real-time performance, which is critical for this class of tasks. 
The contributions of this paper can be summarized as follows:

\begin{itemize}
\item A normal flow-based motion segmentation framework,  which enables accurate identification of IMOs in the scene without relying on prior knowledge. 
This is achieved by formulating the motion segmentation task as an energy minimization problem solved via graph cuts, where normal flow clustering and motion model fitting are performed iteratively.
\item A normal flow-based motion model initialization and fitting method, which enables fast estimation of motion models for IMOs using only a limited number of candidate models. 
This substantially reduces computational complexity and enables real-time system performance.
% A novel normal-flow-based real-time motion segmentation framework. 
% Benefiting from accurate semi-dense normal flow estimation results, we formulate the motion segmentation task as an energy minimization problem using graph cuts, optimize it iteratively with multi-model fitting, and leverage normal flow for fast model initialization, thereby achieving real-time motion segmentation.
\item A comprehensive evaluation on multiple public datasets demonstrates the efficiency and accuracy of the proposed motion segmentation framework.
Our code will be open-sourced to facilitate further research in this field.
\end{itemize}

\emph{Outline}: The rest of this paper is organized as follows.
We begin by reviewing the event-based motion segmentation problem and the concept of event-based normal flow (Sec.~\ref{sec:related}).
Next, we present our method, providing a detailed discussion that highlights the contributions outlined in Sec.~\ref{sec:method}. 
The experimental evaluation is then provided in Sec.~\ref{sec:experiment}, followed by a conclusion in Sec.~\ref{sec:conclusion}.

\section{Related Work}
\label{sec:related}
In this section, we provide a detailed survey of recent progress on both event-based segmentation (Sec.\ref{subsec: ev-motion-seg}) and normal flow (Sec.\ref{subsec:ev-normal-flow}).
% Event-based motion segmentation involves partitioning event streams into distinct clusters, each corresponding to either the background or an independent moving object. 
% The introduction of event cameras proves particularly beneficial in high-dynamic environments, enabling the capture of fast movements without the motion blur common in traditional cameras. 
% However, the camera's ego-motion induces additional complexity, as it generates events across the entire image plane~\cite{glover2016event}\cite{mitrokhin2018iros}\cite{Stoffregen19iccv}, complicating the segmentation process~\cite{gallego2020PAMI}.}
% The key challenge in addressing this problem is determining how to effectively assess data association corresponding to each independently moving object (IMO).  
% One influential approach is motion compensation~\cite{gallego2018cvpr}\cite{gallego2019CVPR}, in which events are warped according to geometric models, and the goodness of fitting is evaluated using the variance of the Image of Warped Events (IWE).
\subsection{Event-based motion segmentation}
\label{subsec: ev-motion-seg}
% Motion segmentation with event cameras involves partitioning event streams into distinct clusters, each corresponding to either the background or an independently moving object. 
% Yet the introduction of event cameras proves particularly beneficial in high-dynamic environments, the camera's ego-motion induces additional complexity, as it generates events across the entire image plane~\cite{glover2016event}\cite{mitrokhin2018iros}\cite{Stoffregen19iccv}, complicating the segmentation process~\cite{gallego2020PAMI}.
A naive idea~\cite{mitrokhin2018iros} of identifying IMOs is to directly apply global motion compensation~\cite{gallego2018cvpr} to the event stream using parametric models. Then clusters of events that deviate from the motion model (residual motion) are labeled as IMOs. 
Compared to~\cite{mitrokhin2018iros}, ~\cite{Stoffregen19iccv} improves event-wise segmentation by jointly estimating event-object associations and object motion parameters, achieving significantly better performance on datasets while distinguishing subtle relative motion differences between clusters. 
Despite these improvements, the choice of motion models remains critical, as incorrect or inadequate model choices can still lead to segmentation failures. 
Parameshwara \etal~\cite{parameshwara2020moms, parameshwara20210} advance event clustering by incorporating motion compensation alongside feature tracking, cluster splitting, and merging. 
This integration strengthens robustness over extended sequences but faces challenges in merging spatially distant clusters exhibiting similar motion. 

Zhou \etal~\cite{zhou2021emsgc} further discuss the clustering nature of the segmentation problem and designed a space-time graph representation to solve this problem in a joint optimization manner.
They reformulate the expectation-maximization (EM)~\cite{meng1997em-algorithm} approach into a minimization framework using negative IWE, ensuring compatibility with graph-cut algorithms.
% \textcolor{cyan}{TODO: Introduce EMSGC again}
Inspired by this framework, Lu \etal~\cite{lu2021iros} develop a cascaded two-stage multi-model fitting scheme that addresses explicitly the cluster merging problem.
Mitrokhin \etal~\cite{mitrokhin2020learning} introduce a graph convolutional network for scene motion segmentation, which effectively learns to perform foreground-background segmentation tasks. 
Using Inertial Measurement Unit (IMU) measurements as camera motion,~\cite{zhou2024icra_jstr} projects events to a reference time and extracts an assumed columnar structure from the point cloud to identify moving objects. 
However, the reliance on strong shape assumptions restricts its applicability across diverse real-world datasets.
A multi-modal model for instance-level moving object segmentation is proposed in \cite{wan2025ijcv}, yet its reliance on RGB images limits applicability in most scenarios where such data is unavailable.

% However, the reliance on strong shape assumptions restricts applicability across diverse real-world datasets.

% In addition to geometric model fitting via motion compensation, learning-based methods are also applied to this task. 

% Geometric self-labeling is applied in~\cite{wang2024eccv} to learn IMO labels and it can run at realtime.
\subsection{Event-based normal flow}
\label{subsec:ev-normal-flow}
Characterizing the relative motion between the camera and dynamic scenes, optical flow\cite{horn1981determining} is a representation of the movement of pixels over time, encoding the direction and velocity of motion.
However, since event cameras capture only pixels where brightness changes occur, they can recover only partial components of optical flow—specifically, the normal flow—through local event data~\cite{benosman2013event, akolkar2020pami}.
Estimation of relative camera motion from normal flow observations has been studied in \cite{parameshwara2022diffposenet}.
For event cameras, normal flow has been used for drones' course estimation \cite{dinaux2021faith} and obstacle avoidance \cite{Clady14fns}. 
Ren \emph{et al.}~\cite{ren2024eccv} propose that the normal flow constraint can serve as an alternative to full optical flow, addressing motion and structure problems in static scenes.
Similar usage of event-based normal flow has also been witnessed for state estimation\cite{lu2024eviv, lu2025tro} and autonomous driving\cite{li2024eccv}.
Nevertheless, their method for computing normal flow is relatively simplistic, and the sparse normal flow they obtain lacks robustness to noise, limiting its effectiveness in dynamic and noisy environments.

The computation of visual flow from event data has been witnessed from both geometric~\cite{benosman2013event, akolkar2020pami, nagata2021sensors} and data-driven~\cite{Wan22TIP, kepple2020eccv, Shiba22eccv} perspectives.
Only recently, VecKM\_Flow~\cite{yuan2024learning} and its real-time variation \cite{yuan2025real} are introduced to directly learn event-wise normal flow from event neighborhoods through a scalable local event encoder named VecKM \cite{pmlr-v235-yuan24b}, offering an improvement in handling dynamic conditions. 
Building on normal flow, \cite{hua2025motion} uses hierarchical clustering for motion segmentation and ego-motion estimation. 
Though achieving fair accuracy, the approach is computationally expensive and nowhere near real-time performance. 

% \yang{TODO: state event-based motion segmentation}
% Identifying Independently Moving Objects (IMOs) without supervision is a key problem in dynamic scene understanding. 
% For event cameras, early literature typically require prior knowledges such as the number or/and shapes of IMOs~\cite{}.
% 

% Inspired by this framework, Lu \etal~\cite{lu2021iros} developed a cascaded two-stage multi-model fitting scheme that specifically addresses cluster merging problem.
% In addition to geometric model fitting via motion compensation, learning-based methods are also applied to this task. 
% Mitrokhin \etal~\cite{mitrokhin2020learning} introduced a Graph Convolutional Network for scene motion segmentation, which effectively learns to perform foreground-background segmentation tasks. 
% Geometric self-labeling is applied in~\cite{wang2024eccv} to learn IMO labels and it can run at realtime.
% Wan \etal~\cite{wan2025ijcv} developed a multi-modal model for instance-level moving object segmentation, yet its reliance on RGB images limits applicability in most scenarios where such data is unavailable.

% To address these limitations, Zhou \emph{et al.} proposed EMSGC framework, modeling event-based segmentation problem as a 

\begin{figure*}[t]
    \vspace{1em}
    \centering
    \includegraphics[width=\linewidth]{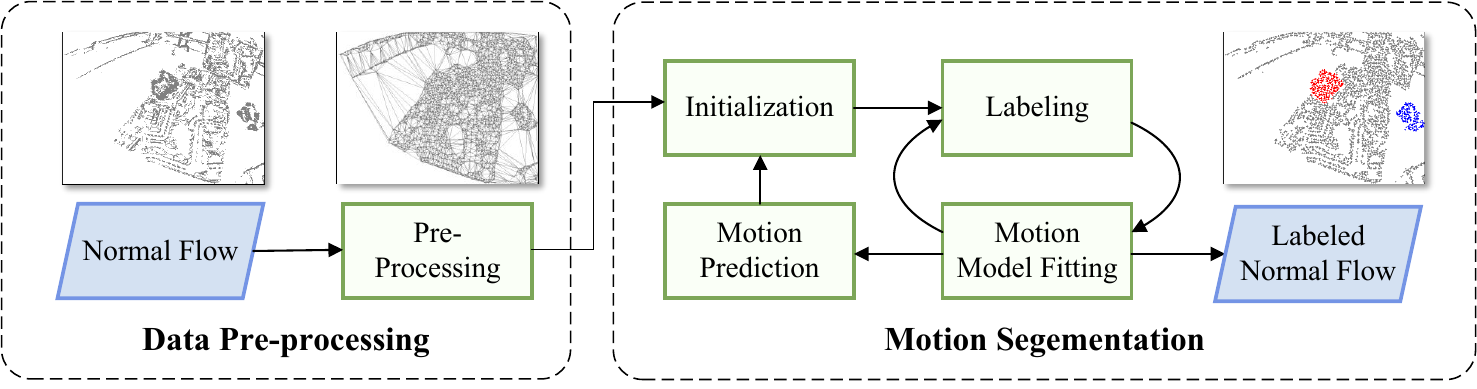}
    \caption{\label{fig:system_overview}
    \emph{Flowchart of the proposed system.} 
    The proposed system comprises two independently operating modules. The data pre-processing module downsamples the input dense normal flow and constructs a spatial graph via Delaunay triangulation~\cite{shewchuk2009general}. 
    The motion segmentation module iteratively alternates between normal flow clustering (Labeling) and motion model fitting to segment the normal flow  associated with IMOs.
    }
    \vspace{-1.5em}
\end{figure*}
\flushbottom

\section{Methodology}
\label{sec:method}
We detail our method in this section. 
First, we provide an overview of the proposed system (Sec.~\ref{subsec: sys oveview}).
% Second, we discuss how the motion segmentation problem is formulated as an energy minimization optimization problem solvable via graph cuts. (Sec.~\ref{subsec: problem statement}). 
Second, we present preliminaries on problem formulation, including normal flow and motion model fitting based on normal flow (Sec.~\ref{subsec: problem statement}).
Third, we discuss how to formulate motion segmentation as an energy minimization problem solved via iterative normal flow clustering and motion model fitting (Sec.~\ref{subsec: solution}).
% Third, we discuss how iterative normal flow clustering and motion model fitting can be performed to yield accurate motion segmentation results (Sec.~\ref{subsec: solution}).
Finally, we introduce the efficient initialization strategy employed in the proposed system (Sec.~\ref{subsec: initialization}).

\subsection{System Overview}
\label{subsec: sys oveview}
The goal of our motion segmentation framework is to find a labeling function $L(
\bfn):\mathcal{N} \rightarrow \mathcal{L} = \{1, \ldots,N\}$, which assigns to the input dense normal flow $\bfn \in\mathcal{N}$ different labels $l \in\mathcal{L}$ based on the motion models $\mathcal{M}=\{\mathbf{m}_{1}, \ldots, \mathbf{m}_{N} \}$, thereby determining the positions and number of IMOs in the scene, with the overall operational logic being analogous to that of EMSGC~\cite{zhou2021emsgc}.
% The overall operational logic of our system is analogous to EMSGC~\cite{zhou2021emsgc}.
Specifically, the proposed system comprises two independently operating modules and takes as input the dense normal flow generated by VecKM\_flow~\cite{yuan2024learning}, as illustrated in Fig.~\ref{fig:system_overview}.
In the data pre-processing module, the normal flow is downsampled at fixed time intervals, followed by the construction of a spatial graph via Delaunay triangulation~\cite{shewchuk2009general}, which is then fed into the motion segmentation module.
In the motion segmentation module, an initial set of candidate motion models is first generated through an initialization step (Sec.~\ref{subsec: initialization}).
Subsequently, an iterative process of normal flow clustering and motion model fitting is performed until the motion models converge, upon which the segmented normal flow is output (Sec.~\ref{subsec: solution}). 
Notably, after motion prediction, regions containing IMOs are employed to initialize the motion models for the next segmentation. 
This effectively reduces the number of candidate motion models required, thereby enhancing system efficiency.

\subsection{Problem Statement Preliminary}
\label{subsec: problem statement}
% In this paper, event-based motion segmentation problem is formulated as an energy function due to its clustering nature and solved in an optimization-based manner.
% The goal of our motion segmentation framework is to find a labeling function $L(
% \bfn):\mathcal{N} \rightarrow \mathcal{L} = \{1, \ldots,N\}$, which assigns the input semi-dense normal flow $\bfn \in\mathcal{N}$ to different labels $l \in\mathcal{L}$ based on the motion models $\mathcal{M}=\{\mathbf{m}_{1}, \ldots, \mathbf{m}_{N} \}$, thereby determining the positions and number of IMOs in the scene.
% The goal of our motion segmentation framework is to assign the input semi-dense normal flow $\mathcal{N}$ to different labels $L:\mathcal{N} \rightarrow \mathcal{L} = \{1, \ldots,N\}$ based on the motion models $\mathcal{M}=\{\mathbf{m}_{1}, \ldots, \mathbf{m}_{N} \}$, thereby determining the positions and number of IMOs in the scene.
% The goal of our motion segmentation framework is to cluster input semi-dense normal flow $\mathcal{N}$ based on motion models, thereby determining the positions and number of Independently Moving Objects (IMOs) in the scene.
% In this section, we first disclose geometric model fitting on event data and how to evaluate the goodness of it.
% Subsequently, we provide a formulation that describes our motion segmentation framework.
Before detailing our method, we first revisit two essential concepts as preliminaries, including normal flow and motion model fitting based on normal flow.
\subsubsection{Event-based Normal Flow and Constraint}
Event cameras capture motion primarily perpendicular to edges, providing only partial observations of optical flow $\bfu$ in the direction of the local image gradient ($\nabla I$). 
This is referred to as normal flow $\bfn = \bfu_{\perp}$. Using the definition of the time surface, i.e. $\Sigma_e(\bfx): \bfx \rightarrow t$, which maps 2D pixel coordinates to 1D timestamps, the  normal in event space is defined as:
\begin{equation}
\label{eq:normal_flow_computtion_from_events}
\bfn(\bfx) = \frac{\nabla \Sigma_e(\bfx)}{\left\| \nabla \Sigma_e(\bfx)\right\|^2_2}.
\end{equation}
Here, $\nabla \Sigma_e(\bfx)$ represents the spatial gradient  of the time surface at $\bfx$.
Leveraging the relationship between both flows, Ren \textit{et al.}\cite{ren2024eccv} demonstrate that event-based normal flow can serve as an alternative to full optical flow by introducing a constraint, as expressed in the following equation, referred to as the normal flow constraint:
\begin{equation}
\label{eq:geometric_error}
\bfn(\bfx)^\top \bfu(\bfx; \bparams) - \|\bfn(\bfx)\|_2^2 \doteq 0.
\end{equation}
The normal flow constraint can effectively replace optical flow in solving various geometric model fitting problems (with parameters denoted as $\bparams$).
It provides a computationally efficient alternative, especially in dynamic scenes, and enhances robustness in tasks like motion estimation and scene geometry, where traditional flow estimation may be challenging or prone to errors.

% Decomposing the optical flow $\bfu = \bfu_{\perp} + \bfu_{\parallel}$ into its normal and parallel components to the local image gradient ($\nabla I$), 
% the normal flow is $\bfn = \bfu_{\perp}$, 
% and the dot-product with $\bfn$ gives
% \begin{equation}
% \label{eq:normalflow:dotprod}
% \bfn^\top \bfu  = \|\bfn\|^2, \quad\text{ or equivalently }\quad (\bfu- \bfn)\cdot \bfn = 0.
% \end{equation}
\subsubsection{Motion model fitting on normal flow}
% Given that full flow can not be observed from local events, the normal flow could serve as its alternative in a lot of geometric model fitting tasks~\cite{ren2024eccv}.
% While effective, these motion field-based models typically either rely on depth prior from other sensors or planar structure assumption.
% To expand normal flow constraint to more general case, we involve the affine model into it.
Similar to motion model fitting via the contrast maximization framework~\cite{gallego2018cvpr} for event data, normal flow derived from local event observations can be employed for geometric model fitting tasks~\cite{ren2024eccv}.
% While effective, most motion models including motion field~\cite{longuet1980interpretation} and continuous homography~\cite{Ma04book} typically require accurate depth priors or planar structure assumptions, which are typically absent in motion segmentation tasks. 
Assuming objects are not very large, usually motion models that implicitly model the geometry of the scene are assumed for segmentation.
To maintain generality, we employ the affine motion model as previous works do~\cite{mitrokhin2018iros, zhou2021emsgc}.
The basic 4-parameter affine model can be written as 
\begin{equation}
\label{eq:affine model}
    \begin{bmatrix}
x'\\ y' \\ 1 

\end{bmatrix}
= 
\begin{bmatrix}
\rho\cos\theta &-\rho\sin\theta & t_x \\  
\rho\sin\theta &\rho\cos\theta & t_y \\  
0 & 0 & 1
\end{bmatrix}
\begin{bmatrix}
x\\ y \\ 1 

\end{bmatrix},
\end{equation}
where $x'$ and $y'$ represent the pixel coordinates after the transformation, and $(\rho, \theta, t_x, t_y)$ are the parameters defining the affine transformation.
Then, the homogeneous coordinate of optical flow $\hat{\textbf{u}}$ can be denoted as:
% \begin{bmatrix}
% x'\\ y' \\ 1 

% \end{bmatrix}
% - 
% \begin{bmatrix}
% x \\ y \\ 1 
% \end{bmatrix}
% =
\begin{equation}
   \hat{\textbf{u}} =     
(\begin{bmatrix}
\rho\cos\theta &-\rho\sin\theta & t_x \\  
\rho\sin\theta &\rho\cos\theta & t_y \\  
0 & 0 & 1
\end{bmatrix}-I)
\begin{bmatrix}
x\\ y \\ 1 
\end{bmatrix},
\end{equation}
Combining this homogeneous representation with normal flow constraint in Eq.~\ref{eq:geometric_error}, we derive
\begin{equation}
    {\bfn}^\top  
    \begin{bmatrix}
        \rho\cos\theta - 1 &-\rho\sin\theta & t_x \\
        \rho\sin\theta   & \rho\cos\theta- 1   & t_y
    \end{bmatrix}
    \begin{bmatrix}
        x\\ y \\ 1 
    \end{bmatrix}
    = \|{\bfn}\|_2^2 ,\\
    \label{eq: fitting error}
\end{equation}
which can be used to assess the goodness of fitting.
% \begin{equation}
%     \bfn(\bfx)^\top     
%   \bfu(\bfx, \bparams) - ||\bfn(\bfx)||^2
% \label{eq: data term error}
% \end{equation}

% Specifically, we 
% \subsubsection{Normal Flow From Event Data}
% Due to the aperture problem, local observation could only recover the partial component of optical flow.
% Thus, 

% VecKMFlow~\cite{yuan2024learningnormalflowdirectly}
% \subsubsection{Event-based Motion Segmentation}
% Building upon CMax~\cite{Gallego18cvpr} framework, EMSGC~\cite{zhou2021emsgc}

\subsection{Solution for Motion Segmentation}
\label{subsec: solution}

% But compared to EMSGC\cite{zhou2021emsgc}, the energy function differs from the data term.

As illustrated in Sec.~\ref{subsec: sys oveview}, we solve the problem by alternately minimizing the two subproblems of normal flow labeling and motion model fitting of each labeled cluster to optimize $L$ and $\mathcal{M}$, respectively.
Following a similar strategy as employed in EMSGC~\cite{zhou2021emsgc}, the motion segmentation task is formulated as an energy minimization problem that involves fitting multiple motion models:
\begin{equation}
\label{eq:energy function}
    E(L, \mathcal{M}) \dot{=} E_{\mathrm{D}}(L, \mathcal{M}) + \lambda_{\mathrm{P}} E_{\mathrm{P}}(L) + \lambda_{\mathrm{M}} E_{\mathrm{M}}(L),
\end{equation}
% where the optimization variables $L$ and $\mathcal{M}$ denote the labeling configuration and clustered motion models, respectively.
% where $E_{\mathrm{D}}$ denotes the data term, while $E_{\mathrm{P}}$ and $E_{\mathrm{M}}$ represent the smoothness term~\cite{potts1952some} and label cost term~\cite{delong2012fast}, respectively.
% where the labeling function $L:\mathcal{N} \rightarrow \mathcal{L} = \{1, \ldots,N\}$ assigns each normal flow to a label (with each label corresponding to an IMO), and $\mathcal{M}=\{\mathbf{m}_{1}, \ldots, \mathbf{m}_{N} \}$ denotes the associated motion model parameters.
where $E_D(L, \mathcal{M})$ is the data term, representing the model fitting error, while $E_P(L)$ and $E_M(L)$ correspond to the smoothness term~\cite{potts1952some} and label cost term~\cite{delong2012fast}, respectively. 
The weights $\lambda$ govern the contribution of each term.

We begin by introducing the labeling process. 
In the pre-processing step, the input event-wise normal flow is constructed into a 2D spatial graph via Delaunay triangulation~\cite{shewchuk2009general}, establishing adjacency relationships. 
Once the candidate motion models $\mathcal{M}$ obtained from initialization are fixed, Eq.~\ref{eq:energy function} is transformed into an MRF problem~\cite{russell2011energy} that can be solved using the alpha-expansion algorithm~\cite{boykov2001fast}, thereby yielding the labeling function $L$.

As for motion model fitting, after fixing the labels $L$, data term $E_{\mathrm{D}}( \mathcal{M})$ is the only remaining component in Eq.~\ref{eq:energy function}. 
Thus, the model fitting process could be resolved by either a linear or a nonlinear optimization method.
% Depending on the formulation of model fitting error, there are two approaches for optimizing the motion models.
For the linear solution, according to Eq.~\ref{eq: fitting error}, the fitting error of each normal flow observation can be reformulated as:
\begin{equation}
    \begin{bmatrix}
        n_x x, n_x y, n_x,
        n_y x, n_y y, n_y
    \end{bmatrix}
    \mathbf{m} = || \bfn||_2^2,
\end{equation}
where $n_x$, $n_y$ denote the component along the x and the y axes of normal flow, and $\mathbf{m}$ is the vector representation of affine transformation:
\begin{equation}
\mathbf{m} = 
        \begin{bmatrix}
        \rho\cos\theta - 1 &-\rho\sin\theta & t_x 
       & \rho\sin\theta   & \rho\cos\theta- 1   & t_y
    \end{bmatrix}^\top.
\end{equation}
By stacking multiple normal flow observations from the same label, we can construct a linear equation:
\begin{equation}
    \begin{pmatrix}
        n_{1x} x, n_{1x} y, n_{1x},
        n_{1y} x, n_{1y} y, n_{1y} \\
        \vdots \\
        n_{kx} x, n_{kx} y, n_{kx},
        n_{ky} x, n_{ky} y, n_{ky}
    \end{pmatrix}
    \mathbf{m} =  \begin{pmatrix}
        ||\bfn_1||_2^2 \\
        \vdots \\
        ||\bfn_k||_2^2 
    \end{pmatrix}.
\end{equation}
After solving the above overdetermined system, the affine model parameters can be decoupled by
\begin{equation}
    \begin{cases}
        \rho = \sqrt{((m_1+m_5)/2+1)^2 + ((m_2-m_4)/2)^2 } \\ 
        \theta = \arcsin{(\frac{m_4-m_2}{2\rho})}\\
        v_x = m_3\\
        v_y = m_6
    \end{cases},
\end{equation}
where $m_i$ is the $i$th entry of $\mathbf{m}$.
Although the linear solution is efficient, the presence of outliers can cause instability when the number of normal flow values is large. 
Therefore, the proposed system typically formulates the model fitting error (Eq.~\ref{eq: fitting error}) as a least squares problem:
\begin{equation}
 E_\mathrm{D}(\mathcal{M}) =\Sigma_{\bfx_k \in \varepsilon_l}(\bfn(\bfx_k)^\top     
  \bfu(\bfx_k, \bparams) - ||\bfn(\bfx_k)||_2^2)^2 .
\end{equation}
Then, the affine model parameters $\bparams = (\rho, \theta, t_x, t_y) $ are optimized using the Levenberg-Marquardt algorithm.

% Building upon the normal flow constraint\cite{ren2024eccv}, the data term is formulated as

\begin{figure}[t]
	\centering
	\subfigure[Result at $t-1$]{\gframe{\includegraphics[width=0.15\textwidth]{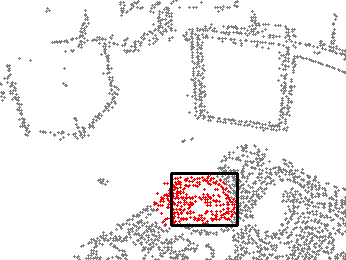}}}\hspace{1pt}
    \subfigure[Input at $t$]{\gframe{\includegraphics[width=0.15\textwidth]{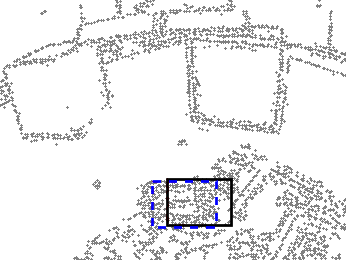}}}\hspace{1pt}
	\subfigure[Result at $t$]{\gframe{\includegraphics[width=0.15\textwidth]{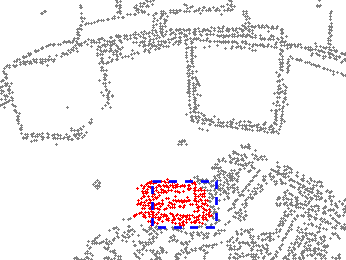}}}
	% \caption{\textit{Different strategies of sampling points.} (a) is the strategy adopted in ESVO that samples 10,000 pixels with the latest timestamp within a certain time interval,  which concentrated in the regions of significant optical flow. (b) is sampling results of ESVO with 5000 pixels. (c) is the 5000 pixels sampled on AA with the highest number of accumulated events, which are more evenly and continuously distributed on the dge structure.}
 \caption{
 \textit{Procedure of motion prediction.}
 {(a)} Segmentation result at $t-1$, with the black solid box indicating the system-generated region containing an IMO.
 {(b)} Normal flow input at $t$, with the blue dashed box indicating the predicted IMO region after motion prediction. 
 Normal flow within this region is used to initialize a candidate motion model.
 {(c)} Segmentation result at $t$, where the IMO primarily resides within the predicted box.}
    \label{fig:initialiaztion}
\vspace{-1em}
\end{figure}
\subsection{Initialization}
\label{subsec: initialization}
% Our initialization strategy deviates fundamentally from existing event-based motion segmentation approaches due to its unique system input requirements.
% Based on the characteristics of normal flow and the temporal continuity of independently moving objects, we design a simple and efficient initialization strategy for motion model labels, which consists of two steps:
We now detail how the optimization procedure is initialized.
Leveraging the characteristics of normal flow and the motion continuity of IMOs, we propose a simple and efficient initialization strategy, which consists of two parts:

\subsubsection{Fast Sampling}
% In the first step, given a normal flow set $\mathcal{N}$ as input, we directly traverse this set to select $\mathit{n}$ normal flows exhibiting distinct directional and magnitude differences. 
% Their direction and magnitude (represented in planar Cartesian coordinates) are assigned as the two-dimensional translation parameters in the affine transformation model. 
% Considering that the scaling and rotation parameters in affine models typically exhibit minimal magnitudes, we initialize them as 1 and 0, respectively, thereby generating $\mathit{n}$ motion model candidates $\mathcal{M}$.
Upon receiving new normal flow in the motion segmentation module, we first sample $n$ instances of normal flow with significantly different translation vectors. 
These translation vectors are then used to initialize the translation components  $(t_x, t_y)$ of the affine motion model as defined in Eq.~\ref{eq:affine model}.
This approach leverages the inherent translational motion information encapsulated within the normal flow, as it constitutes the projection of optical flow onto the gradient direction.
For the scale factor $\rho$ and rotation angle $\theta$, which typically exhibit only minor variations, we directly set them to initial values of 1 and 0, respectively, yielding $n$ complete candidate motion models.

% Given the 4-parameter affine motion model in Eq.~\ref{eq:affine model}, we first consider that the scale factor $\rho$ and rotation angle $\theta$ typically exhibit only minor variations; therefore, they are directly assigned initial values of 1 and 0, respectively.
% Regarding translation components $(t_x, t_y)$, since the normal flow is the projection of optical flow along the gradient direction and inherently contains translational motion information, we sample $n$ instances of normal flow with significantly different translation vector.
% These translation vectors are then sequentially assigned to $t_x$ and $t_y$, yielding $n$ complete candidate motion models.

\subsubsection{Motion Prediction}
After performing motion segmentation at time $t-1$, for each label identified as belonging to an IMO, we first apply region growing to generate a bounding box that encompasses the IMO as much as possible. 
Given the corresponding motion model and the system's operation rate, we then predict the box's position at time $t$, as shown in Fig.~\ref{fig:initialiaztion}.
Before beginning the motion segmentation at time $t$, the normal flow within each predicted box is used to fit a candidate motion model by nonlinear optimization.
The candidate motion models obtained using this method can effectively improve the accuracy of motion segmentation when motion prediction is reliable.
Consequently, when these models are available, the number of models generated by fast sampling is reduced (6 in our implementation when available, and 12 otherwise).

% In the second step, after each motion segmentation, we perform region growing for each of the $\mathit{m}$ labels identified as independent moving objects to obtain minimally enclosing bounding boxes. 
% Given the the system's operation rate, we then predict the approximate spatial regions where these objects will appear in subsequent frames. 
% Nonlinear optimization is applied to the normal flow within these predicted regions to obtain $\mathit{m}$ motion model candidates, also denoted as $\mathcal{M}$. 
% It is noteworthy that if step two successfully predicts the motion models of independently moving objects, the number of normal flow selected in step one is significantly reduced (6 for success and 12 for failure in our implementation), as fewer motion model candidates are still sufficient to accurately identify independently moving objects.

% The efficacy of our initialization strategy largely stems from the inherent richness of motion information encapsulated within the normal flow. 
% In contrast, EMSGC~\cite{zhou2021emsgc} operates directly on raw event data, necessitating multiple subdivision operations on the event stream followed by motion-compensated estimation of up to 85 candidate motion labels under the CMax~\cite{Gallego18CVPR} framework, which is computationally intensive. 
% By leveraging the normal flow, our method rapidly derives accurate motion models through direct parameter assignment and nonlinear optimization, thereby enabling discrimination of independent moving objects with fewer labels and substantially reducing computational overhead.

The proposed initialization strategy can initialize a sufficient number of candidate motion models with minimal computational cost while achieving high-quality motion segmentation results. 
In contrast, EMSGC performs an N-level subdivision operation to divide the event volume and then initializes 85 candidate motion models on the sub-volumes through motion compensation, which results in significantly higher computational complexity
Detailed computational performance can be found in Sec.~\ref{subsec:computational efficiency}.
This highlights the advantage of using normal flow over directly operating on raw event streams.

\def\figWidth{0.18\linewidth} 
\begin{figure*}[t]
    \vspace{1em}
    \centering
    {\small
    \setlength\tabcolsep{2pt} % Spacing between columns
    \begin{tabular}{
	>{\centering\arraybackslash}m{0.3cm} 
	>{\centering\arraybackslash}m{\figWidth} 
	>{\centering\arraybackslash}m{\figWidth}
	>{\centering\arraybackslash}m{\figWidth}
	>{\centering\arraybackslash}m{\figWidth}
	>{\centering\arraybackslash}m{\figWidth}
    }

    % & Scene & Time Surface (left) & Selected points & Inverse depth map & 3D reconstruction \\
    
	\rotatebox{90}{\makecell[c]{\ \emph{Fast Drone\_logo}}} & 
    \gframe{\includegraphics[trim={4px 4px 4px 4px},clip,width=\linewidth]{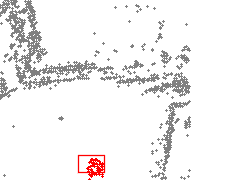}} & 
    \gframe{\includegraphics[trim={4px 4px 4px 4px},clip,width=\linewidth]{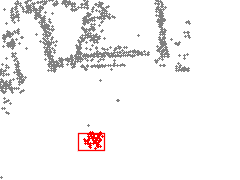}} &
    \gframe{\includegraphics[trim={4px 4px 4px 4px},clip,width=\linewidth]{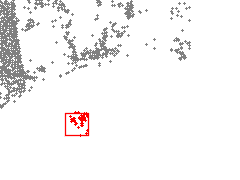}} &
    \gframe{\includegraphics[trim={4px 4px 4px 4px},clip,width=\linewidth]{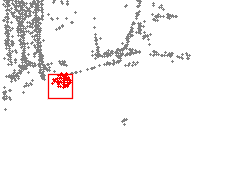}} &
	\gframe{\includegraphics[trim={4px 4px 4px 4px},clip,width=\linewidth]{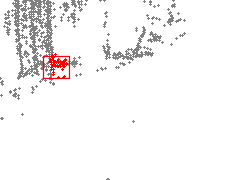}} \\%\addlinespace[-0.3ex]

    \rotatebox{90}{\makecell[c]{\ \emph{Light Variation}}} & 
    \gframe{\includegraphics[trim={4px 4px 4px 4px},clip,width=\linewidth]{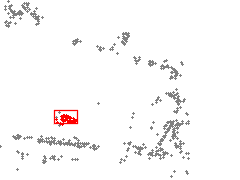}} & 
    \gframe{\includegraphics[trim={4px 4px 4px 4px},clip,width=\linewidth]{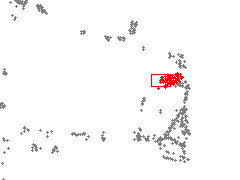}} &
    \gframe{\includegraphics[trim={4px 4px 4px 4px},clip,width=\linewidth]{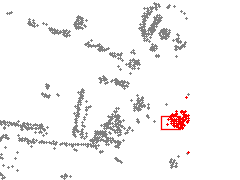}} &
    \gframe{\includegraphics[trim={4px 4px 4px 4px},clip,width=\linewidth]{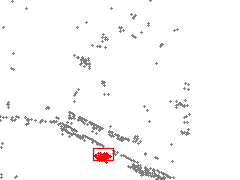}} &
    \gframe{\includegraphics[trim={4px 4px 4px 4px},clip,width=\linewidth]{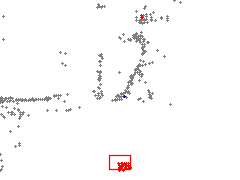}} 
    \\

    \rotatebox{90}{\makecell[c]{\ \emph{Occlusion}}} & 
    \gframe{\includegraphics[trim={4px 4px 4px 4px},clip,width=\linewidth]{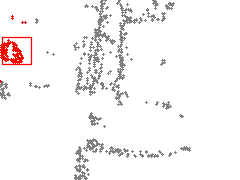}} & 
    \gframe{\includegraphics[trim={4px 4px 4px 4px},clip,width=\linewidth]{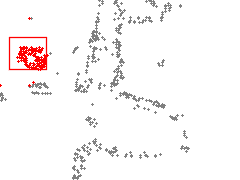}} &
    \gframe{\includegraphics[trim={4px 4px 4px 4px},clip,width=\linewidth]{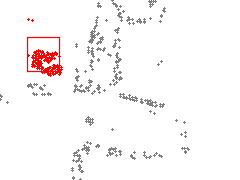}} &
    \gframe{\includegraphics[trim={4px 4px 4px 4px},clip,width=\linewidth]{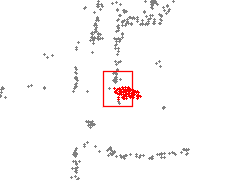}} &
    \gframe{\includegraphics[trim={4px 4px 4px 4px},clip,width=\linewidth]{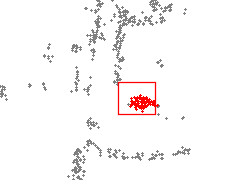}} 
    \\
    
    \rotatebox{90}{\makecell[c]{\ \emph{What is background}}} & 
    \gframe{\includegraphics[trim={4px 4px 4px 4px},clip,width=\linewidth]{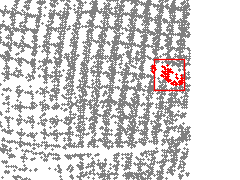}} & 
    \gframe{\includegraphics[trim={4px 4px 4px 4px},clip,width=\linewidth]{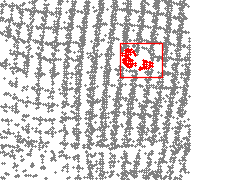}} &
    \gframe{\includegraphics[trim={4px 4px 4px 4px},clip,width=\linewidth]{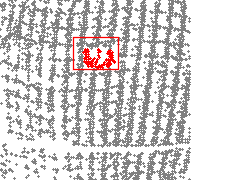}} &
    \gframe{\includegraphics[trim={4px 4px 4px 4px},clip,width=\linewidth]{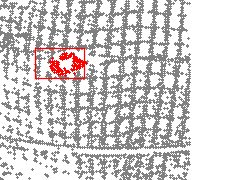}} &
    \gframe{\includegraphics[trim={4px 4px 4px 4px},clip,width=\linewidth]{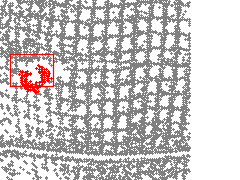}} 
    \\

    \end{tabular}
    }
   \caption{\label{fig:EED}
   \emph{Segmentation results on the EED dataset}~\cite{mitrokhin2018iros}. 
   Time runs from left to right.
   The ground truth bounding boxes are denoted by red rectangles. 
   Since the boxes are manually annotated on the grayscale images, and the timestamps cannot be perfectly aligned with the segmentation results, offsets are witnessed, especially for fast-moving IMOs.}
   \vspace{-1em}
\end{figure*}

\section{Experiment}
\label{sec:experiment}

In this section, we evaluate the proposed system. 
We first describe the experimental datasets and evaluation metrics (Sec.~\ref{subsec:dataset and evaluation}), followed by quantitative and qualitative comparisons against other methods across multiple benchmarks (Sec.~\ref{subsec:evaluation}).
Finally, we analyze the computational efficiency of the proposed system (Sec.~\ref{subsec:computational efficiency}).

\begin{table}[t]
\vspace{2em}
\centering
\caption{\label{tab:camera parameters}Summary of characteristics of the datasets used.}
\begin{adjustbox}{max width=\linewidth}
\renewcommand{\arraystretch}{1.2}
\setlength{\tabcolsep}{1.2mm}{
\begin{tabular}{llccc}
\toprule
\textbf{Dataset} & \textbf{Camera} & \textbf{Resolution} & \textbf{Env.}& \textbf{HDR} \\
\midrule

EED~\cite{mitrokhin2018iros} & DAVIS240 & 240$\times$180 & Indoor & Yes\\
EVIMO~\cite{mitrokhin2019IROS} & DAVIS346 & 346$\times$260 & Indoor & No\\
EMSGC~\cite{zhou2021emsgc} & DAVIS346 & 346$\times$260 & Outdoor & No\\
\bottomrule
\end{tabular}
}
\end{adjustbox}
\vspace{-2em}
\end{table}

\subsection{Datasets and Evaluation Metrics}
\label{subsec:dataset and evaluation}
To evaluate the system performance, we conduct evaluations on three publicly available event-based motion segmentation datasets, with key characteristics detailed in Tab.~\ref{tab:camera parameters}.
% and a self-recorded dataset, 
The Extreme Event Dataset (EED)~\cite{mitrokhin2018iros} is one of the first open-source datasets for IMO detection and tracking. 
Each sequence contains camera ego-motion and IMOs, collected in laboratory environments featuring controlled high dynamic range conditions.
EVIMO~\cite{mitrokhin2019IROS}, also captured in laboratory settings but under better illumination, features sequences with up to three IMOs and provides dense segmentation masks for evaluation.
% Unlike the above datasets captured with handheld cameras, we mount the DAVIS346 on a moving autonomous vehicle, generating higher relative velocities through counter-motion with IMO to demonstrate the potential of event camera for high-speed moving object detection.
Unlike the aforementioned datasets, EMSGC~\cite{zhou2021emsgc} provides outdoor sequences containing non-rigid IMOs (e.g., pedestrians), on which we conduct qualitative comparisons with the EMSGC algorithm.
In addition, we record a small number of sequences in a room equipped with the NOKOV Mars26H Motion Capture system, which are used for computational time analysis.

\textbf{Evaluation Metrics.}
Due to the varying formats of ground-truth (GT) provided by different datasets, we employed two distinct evaluation metrics.
The first metric is detection rate, introduced in \cite{mitrokhin2018iros}, which evaluates bounding box overlap between detected and GT objects. 
A detection is considered successful when it meets the following criteria:
\begin{equation}
    \mathcal{B}_D \cap \mathcal{B}_G > 0.5~~\text{and}~~\mathcal{B}_D \cap \mathcal{B}_G > \mathcal{B}_D \cap \mathcal{\overline{B}}_G,
\end{equation}
where $\mathcal{B}_D$ refers to the estimated convex hull, ${B}_G$ the GT bounding box, and $\overline{\cdot}$ denotes the operation of the set complement.

\global\long\def\figWidth{0.15\linewidth}  % 调整为 0.22\linewidth 来使图片放得下

\begin{figure*}[t]
\vspace{1em}
\centering
\setlength\tabcolsep{3pt}  % 增大列间距，避免图片重叠
\begin{tabular}{>{\centering\arraybackslash}m{0.3cm} 
        >{\centering\arraybackslash}m{\figWidth} 
        >{\centering\arraybackslash}m{\figWidth}
        >{\centering\arraybackslash}m{\figWidth}
        >{\centering\arraybackslash}m{\figWidth}
        >{\centering\arraybackslash}m{\figWidth}
        >{\centering\arraybackslash}m{\figWidth}
}
    
    \rotatebox{90}{\makecell[c]{\ GT}} & 
    \gframe{\includegraphics[trim={4px 4px 4px 4px},clip,width=\linewidth]{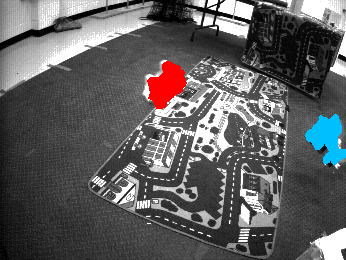}} & 
    \gframe{\includegraphics[trim={4px 4px 4px 4px},clip,width=\linewidth]{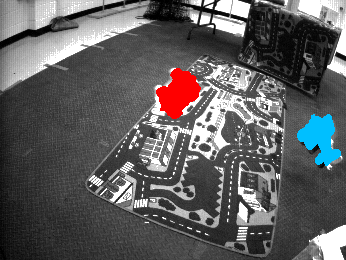}} & 
    \gframe{\includegraphics[trim={4px 4px 4px 4px},clip,width=\linewidth]{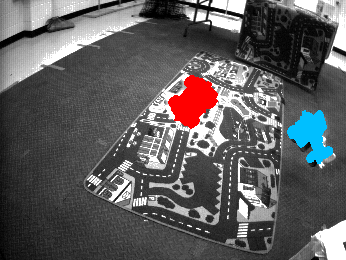}} & 
    \gframe{\includegraphics[trim={4px 4px 4px 4px},clip,width=\linewidth]{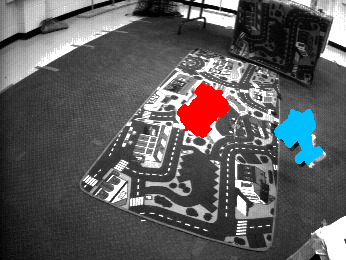}} & 
    \gframe{\includegraphics[trim={4px 4px 4px 4px},clip,width=\linewidth]{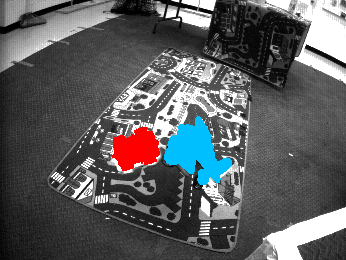}} &
    \gframe{\includegraphics[trim={4px 4px 4px 4px},clip,width=\linewidth]{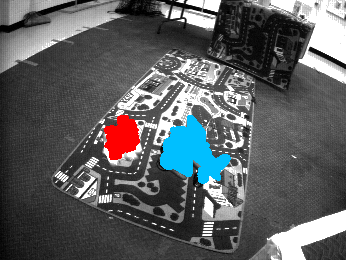}} \\

    \rotatebox{90}{\makecell[c]{\ Ours}} & 
    \gframe{\includegraphics[trim={4px 4px 4px 4px},clip,width=\linewidth]{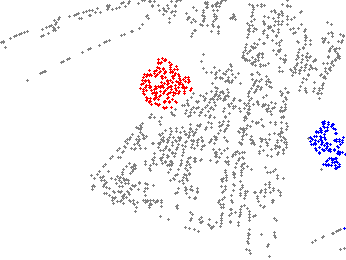}} & 
    \gframe{\includegraphics[trim={4px 4px 4px 4px},clip,width=\linewidth]{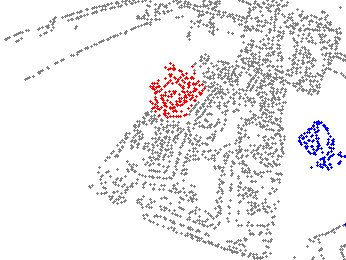}} &
    \gframe{\includegraphics[trim={4px 4px 4px 4px},clip,width=\linewidth]{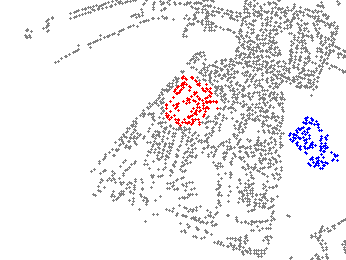}}  &
    \gframe{\includegraphics[trim={4px 4px 4px 4px},clip,width=\linewidth]{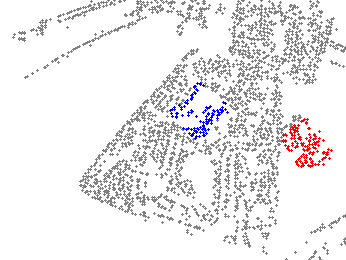}} &
    \gframe{\includegraphics[trim={4px 4px 4px 4px},clip,width=\linewidth]{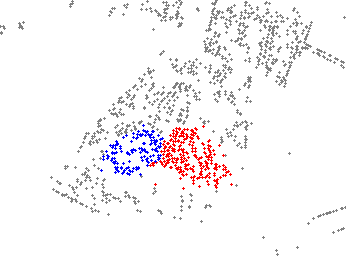}} &
    \gframe{\includegraphics[trim={4px 4px 4px 4px},clip,width=\linewidth]{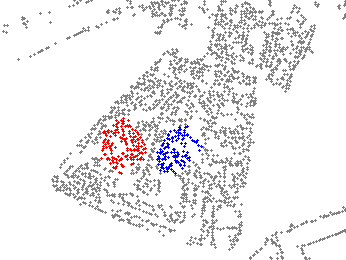}} \\

   \rotatebox{90}{\makecell[c]{\ GT}} & 
    \gframe{\includegraphics[trim={4px 4px 4px 4px},clip,width=\linewidth]{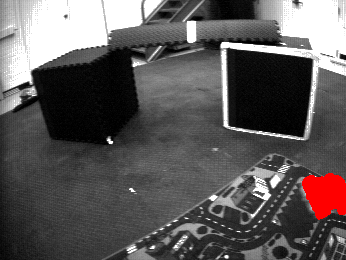}} & 
    \gframe{\includegraphics[trim={4px 4px 4px 4px},clip,width=\linewidth]{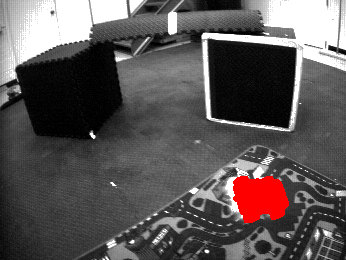}}& 
    \gframe{\includegraphics[trim={4px 4px 4px 4px},clip,width=\linewidth]{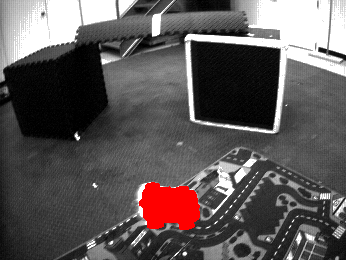}}& 
    \gframe{\includegraphics[trim={4px 4px 4px 4px},clip,width=\linewidth]{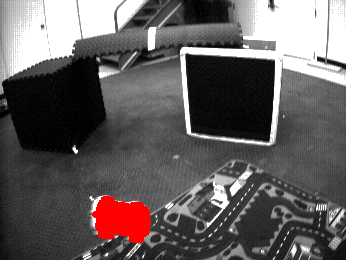}}& 
    \gframe{\includegraphics[trim={4px 4px 4px 4px},clip,width=\linewidth]{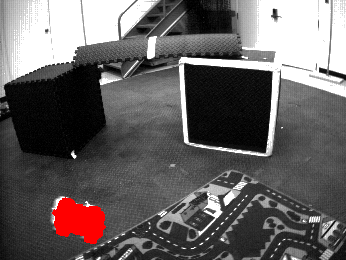}}&
    \gframe{\includegraphics[trim={4px 4px 4px 4px},clip,width=\linewidth]{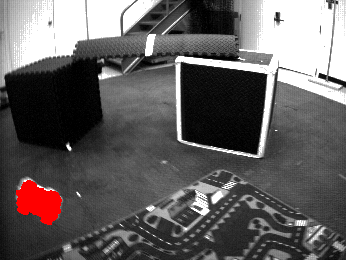}} \\
    
    \rotatebox{90}{\makecell[c]{\ Ours}} & 
    \gframe{\includegraphics[trim={4px 4px 4px 4px},clip,width=\linewidth]{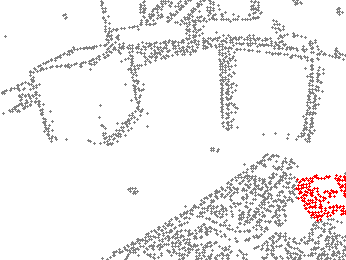}} & 
    \gframe{\includegraphics[trim={4px 4px 4px 4px},clip,width=\linewidth]{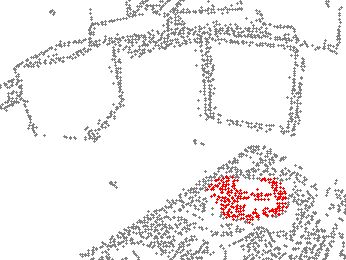}}& 
    \gframe{\includegraphics[trim={4px 4px 4px 4px},clip,width=\linewidth]{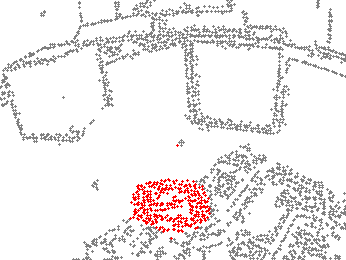}}& 
    \gframe{\includegraphics[trim={4px 4px 4px 4px},clip,width=\linewidth]{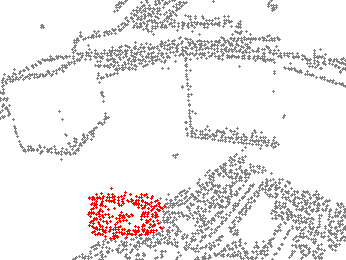}}& 
    \gframe{\includegraphics[trim={4px 4px 4px 4px},clip,width=\linewidth]{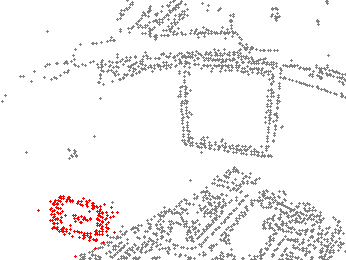}}&
    \gframe{\includegraphics[trim={4px 4px 4px 4px},clip,width=\linewidth]{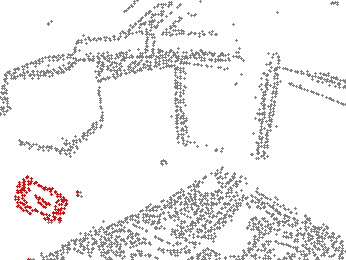}} \\

\end{tabular}

\caption{\textit{Segmentation results on the EVIMO dataset}~\cite{zhou2021emsgc}, \textit{on sequences Table (rows 1-2) and Boxed (rows 3-4).}
Time runs from left to right. 
The grayscale images are used for visualization only. 
The color of each label is determined by the number of normal flow associated with it, meaning that the label color for the same IMO may vary at different time.
The ground truth mask is not perfectly aligned with the IMOs in the grayscale image due to the presence of motion blur.
% Time runs from left to right. 
% Different colors indicate different motion models.
}
\label{fig:EVIMO}
\vspace{0em}
\end{figure*}

\begin{table}[t]
\vspace{-0em}

\begin{center}
\caption{Quantitative evaluation on the EED dataset, with the evaluation metric being detection rate (in \%).}
%The light interferes with the generation of AA, making it difficult to extract high frequency events.
\vspace{0em}
\setlength{\tabcolsep}{2.3mm}{
\renewcommand{\arraystretch}{1.5}
\begin{tabular}{lccc}
\hline
{Sequence} & {EMSMC}~\cite{Stoffregen19iccv} & {EMSGC}~\cite{zhou2021emsgc} & {Ours} \\
\hline
\emph{Fast drone} & \textbf{96.30} &  \textbf{96.30} & \textbf{96.30} \\
\emph{Lighting variation} & 80.51 &  93.51 & \textbf{98.70} \\
\emph{Occlusions} & 92.31 & \textbf{100.00} & \textbf{100.00} \\
\emph{What is background} &100.00 & \textbf{100.00} & \textbf{100.00} \\
\hline
Average & 92.28 & 97.45 & \textbf{98.75}  \\
\hline 
\end{tabular}
}
\label{tab:EED}
\end{center}
% \vspace{-1.5em}
\end{table}
% In motion segmentation, the Intersection over Union (IoU) metric is commonly used to evaluate the accuracy of predicted object masks by comparing them with ground-truth masks. 
The second evaluation metric is Intersection over Union (IoU)~\cite{parameshwara2020moms}, which is calculated as the ratio of the intersection area between the predicted mask $\mathcal{M}_D$ and the GT mask $\mathcal{M}_G$ to the area of their union:
\begin{equation}
\text{IoU} = \frac{\mathcal{M}_D \cap \mathcal{M}_G}{\mathcal{M}_D \cup \mathcal{M}_G},
\end{equation}
A higher IoU indicates better alignment and more accurate segmentation. 
% In this paper, the dense predicted mask is generated by morphology operations from sparse labeled normal flow.
% IoU serves as a crucial evaluation tool in assessing the performance of motion segmentation algorithms, especially when dealing with dynamic environments and complex object interactions.

\begin{table}[t]
% \vspace{-0em}

\begin{center}
\caption{Quantitative evaluation on the EVIMO dataset, with the evaluation metric being IoU.}
%The light interferes with the generation of AA, making it difficult to extract high frequency events.
\vspace{0em}
\setlength{\tabcolsep}{2.3mm}{
\renewcommand{\arraystretch}{1.5}
\begin{tabular}{lcc}
\hline
{Sequence}  & {EMSGC}~\cite{zhou2021emsgc} & {Ours} \\
\hline
\emph{box} & 0.30 & \textbf{0.56} \\
\emph{table} & 0.46 & \textbf{0.55} \\
\hline
Average & 0.38 & \textbf{0.55}\\
\hline 
\end{tabular}
}
\label{tab:EVIMO}
\end{center}
\vspace{-6em}
\end{table}

\subsection{Quantitative and Qualitative Evaluation}
\label{subsec:evaluation}
% We first evaluate the proposed system on the EED dataset~\cite{mitrokhin2018iros}, with the quantitative comparison results shown in Tab.~\ref{tab:EED}. 
We first evaluate our system on the EED dataset~\cite{mitrokhin2018iros}, which covers four extreme scenarios: high-speed motion, illumination changes, occlusions, and moving behind background nets. 
As shown in Tab.~\ref{tab:EED}, our method outperforms two classic motion-compensation-based algorithms, EMSMC~\cite{Stoffregen19iccv} and EMSGC~\cite{zhou2021emsgc}.
The qualitative results can be found in Fig.~\ref{fig:EED}, where the GT bounding boxes are highlighted with red rectangles.
Notably, despite the emergence of several event-based motion segmentation frameworks in recent years, we are unable to compare with these methods due to the use of different datasets and evaluation metrics, as well as the closed-source code.

The second dataset is EVIMO~\cite{mitrokhin2019IROS}, where our method achieves significantly better segmentation results than EMSGC, as shown in Tab.~\ref{tab:EVIMO}. 
The relatively low values stem from two factors: temporal misalignment between our segmentation results and the ground truth timestamps, and the sparser nature of our segmentation outputs (Fig.~\ref{fig:EVIMO}), which introduces errors during mask generation.
The difference in EMSGC's results compared to those in its manuscript is attributed to the different way we used to generate the masks.

Finally, we conduct a qualitative comparison with EMSGC~\cite{zhou2021emsgc} on the outdoor sequence recorded by its authors, as shown in Fig.~\ref{fig:EMS}. 
EMSGC’s initialization strategy causes fragmentation when segmenting non-rigid IMOs, erroneously assigning events from a single IMO to multiple motion models.
In contrast, our method achieves more accurate and consistent segmentation. 
The visually sparser appearance of our results is due to downsampling and a shorter interval.

\global\long\def\figWidth{0.3\linewidth} 
\begin{figure}[t]
\vspace{1em}
\centering
\setlength\tabcolsep{2pt}
\begin{tabular}{>{\centering\arraybackslash}m{0.3cm}
                >{\centering\arraybackslash}m{\figWidth} 
	        >{\centering\arraybackslash}m{\figWidth}
	           >{\centering\arraybackslash}m{\figWidth}}
    
			\rotatebox{90}{\makecell[c]{\ EMSGC}} & 
    \gframe{\includegraphics[trim={4px 4px 4px 4px},clip,width=\linewidth]{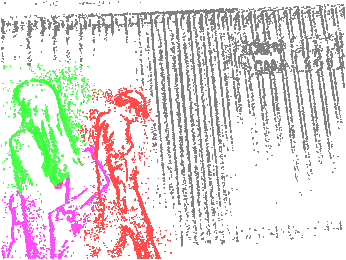}} & 
    \gframe{\includegraphics[trim={4px 4px 4px 4px},clip,width=\linewidth]{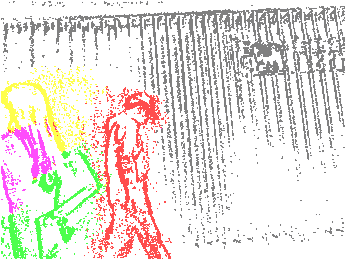}} &
    \gframe{\includegraphics[trim={4px 4px 4px 4px},clip,width=\linewidth]{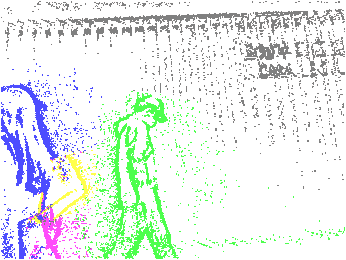}} \\

    \rotatebox{90}{\makecell[c]{\ Ours}} & 
    \gframe{\includegraphics[trim={4px 4px 4px 4px},clip,width=\linewidth]{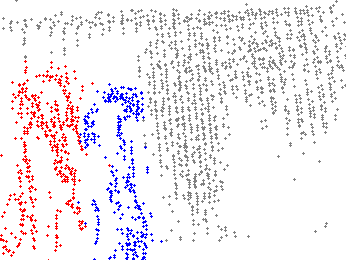}} & 
    \gframe{\includegraphics[trim={4px 4px 4px 4px},clip,width=\linewidth]{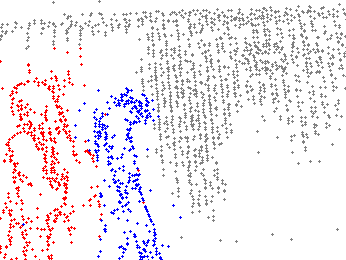}} &
    \gframe{\includegraphics[trim={4px 4px 4px 4px},clip,width=\linewidth]{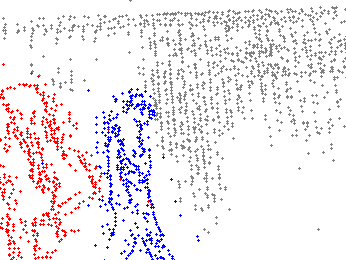}} \\
    
\end{tabular}
% \vspace{2em}
\caption{\textit{Qualitative comparison on the EMSGC dataset}~\cite{zhou2021emsgc}.
Time runs from left to right. 
Different colors indicate different motion models.
}
\label{fig:EMS}
% \vspace{-1.5em}
\end{figure}
\begin{table}[!t]
\vspace{-1em}
\begin{center}
\renewcommand\arraystretch{1.5}
\caption{Computational performance [time: ms].}
\vspace{0em}
\setlength{\tabcolsep}{1.5mm}{
\begin{tabular}{llrr}
\hline
\textbf{Node} & \textbf{Function} & \textbf{EMSGC}~\cite{zhou2021emsgc} & \textbf{Ours}\\

\hline
{Data Pre-processing} & Pre-processing & 87.65 & \textbf{25.50} \\

\hline
\multirow{3}{*}{Motion Segmentation } & Initialization & 5575.97 & \textbf{0.25}  \\
{~} & Labeling \& Fitting & 10892.85 & \textbf{21.73} \\
\cdashline{2-4}
{~} & {Subtotal} & 16468.82 & \textbf{21.98} \\
\hline
\end{tabular}
}
\label{tab:compuational performance}
\vspace{-4.5em}
\end{center}
\end{table}

\subsection{Computational Efficiency}
\label{subsec:computational efficiency}
As shown in Tab.~\ref{tab:compuational performance}, we compare the computational efficiency of EMSGC~\cite{zhou2021emsgc} and the proposed system using a desktop with an Intel Core i7-14700k CPU.
Both systems, implemented in C++ within the ROS framework, are tested on the EVIMO dataset and the self-recorded sequences.
% Benefiting from the characteristics of normal flow, our system achieves significant computational improvements in two key aspects.
Compared to the extremely high computational cost of EMSGC, our system, benefiting from the characteristics of normal flow, significantly reduces computational complexity in two aspects, enabling the entire system to operate in real time at 30 Hz or an even higher rate, which is especially crucial for time-sensitive motion segmentation tasks.

This achievement is attributed to the following two features: in terms of initialization, as mentioned in Sec.~\ref{subsec: initialization}, EMSGC uses the CMax framework~\cite{gallego2018cvpr} for motion model fitting, which is highly sensitive to the initial values of the motion models. 
This necessity to initialize a large number of candidate models to avoid failure, combined with the high computational complexity of CMax's operation on raw event data, results in an initialization time on the order of seconds for EMSGC.
In contrast, the motion information contained in the normal flow allows the proposed system to initialize the motion models simply by traversing the normal flow. 
The high tolerance of the normal flow-based model fitting framework to initial values, along with the motion prediction strategy, significantly reduces the number of required candidate motion models, allowing our system to compress the initialization time to sub-millisecond levels.
Second, the local representation ability of the normal flow significantly reduces the amount of data our system needs, thereby lowering the computational complexity of labeling and motion model fitting. 
The reduction stems from two aspects: spatially, our system utilizes only one normal flow per neighborhood instead of all events; temporally, our system achieves accurate motion segmentation using data over shorter time intervals. 
For instance, on the EVIMO dataset, our system requires only 10 ms or even shorter normal flow for segmentation, whereas EMSGC relies on 50 ms of events. 
With further improvements, there is still significant potential for accelerating our system.

\section{Conclusion}
\label{sec:conclusion}
We propose a real-time motion segmentation system with event-based normal flow. 
The proposed system is built upon EMSGC~\cite{zhou2021emsgc}, an event-based motion segmentation framework, which formulates the problem as an energy minimization optimization problem and solves it using graph cuts. 
To address the computational limitation in EMSGC, we introduce event-based normal flow and, based on its characteristics, propose a novel motion model initialization and fitting method, which enables the proposed system to efficiently estimate the motion models of IMOs with only a limited number of candidate models. 
This significantly reduces the system's computational complexity, enabling real-time operation and representing a significant step forward in event-based motion segmentation for practical applications. 
Extensive evaluations on three publicly available datasets demonstrate the efficiency of our system and its better segmentation performance compared to EMSGC.
Despite these strengths, the current framework's reliance on high-quality normal flow may limit its robustness under extreme conditions. We believe future research could address this by integrating multi-scale flow features or learning-based priors to improve system reliability.
Furthermore, extending the system to incorporate more complex motion models could improve its performance on non-rigid or deformable objects.

% % \addtolength{\textheight}{-12cm}   % This command serves to balance the column lengths
%                                   % on the last page of the document manually. It shortens
%                                   % the textheight of the last page by a suitable amount.
%                                   % This command does not take effect until the next page
%                                   % so it should come on the page before the last. Make
%                                   % sure that you do not shorten the textheight too much.

\bibliographystyle{IEEEtran} % local file, sorted in order of appearance
\bibliography{myBib}
\end{document}